\documentclass[fleqn,10pt]{wlscirep}
\usepackage[utf8]{inputenc}
\usepackage[T1]{fontenc}

\usepackage{hyperref}
\usepackage{amsmath}
\usepackage{algorithm}
\usepackage{algpseudocode}
\usepackage{longtable} % Required for long tables
\usepackage{booktabs} % For better table rules
\usepackage{lipsum}   % For generating dummy text (remove in your actual document)
\usepackage{geometry} % Adjust page margins (optional)

\usepackage{longtable}
\usepackage{booktabs}
\usepackage{array}
\usepackage{geometry}
\usepackage{makecell}
\usepackage{makecell}
\usepackage{graphicx}
\usepackage{xcolor}

\usepackage{longtable}
\usepackage{booktabs}

\title{NeuroSym-BioCAT: Leveraging Neuro-Symbolic Methods for Biomedical Scholarly Document Categorization and Question Answering}

\author[1]{Parvez Zamil}
\author[2,*]{Gollam Rabby}
\author[3]{Md. Sadekur Rahman}
\author[1,2]{Sören Auer}
\affil[1]{TIB—Leibniz Information Centre for Science and Technology, Hannover, Germany}
\affil[2]{L3S Research Center, Leibniz University Hannover, Hannover, Germany}
\affil[3]{Daffodil International University, Bangladesh}

\affil[*]{gollam.rabby@l3s.de}


\begin{abstract}
The growing volume of biomedical scholarly document abstracts presents an increasing challenge in efficiently retrieving accurate and relevant information. To address this, we introduce a novel approach that integrates an optimized topic modelling framework, OVB-LDA, with the BI-POP CMA-ES optimization technique for enhanced scholarly document abstract categorization. Complementing this, we employ the distilled MiniLM model, fine-tuned on domain-specific data, for high-precision answer extraction. Our approach is evaluated across three configurations: scholarly document abstract retrieval, gold-standard scholarly documents abstract, and gold-standard snippets, consistently outperforming established methods such as RYGH and bio-answer finder. Notably, we demonstrate that extracting answers from scholarly documents abstracts alone can yield high accuracy, underscoring the sufficiency of abstracts for many biomedical queries. Despite its compact size, MiniLM exhibits competitive performance, challenging the prevailing notion that only large, resource-intensive models can handle such complex tasks. Our results, validated across various question types and evaluation batches, highlight the robustness and adaptability of our method in real-world biomedical applications. While our approach shows promise, we identify challenges in handling complex list-type questions and inconsistencies in evaluation metrics. Future work will focus on refining the topic model with more extensive domain-specific datasets, further optimizing MiniLM and utilizing large language models (LLM) to improve both precision and efficiency in biomedical question answering.
\end{abstract}

\begin{document}

\flushbottom
\maketitle
\thispagestyle{empty}

%\noindent Please note: Abbreviations should be introduced at the first mention in the main text – no abbreviations lists. Suggested structure of main text (not enforced) is provided below.

\section*{Introduction}

    With around 2.5 million new research contributions every year~\cite{d2020Improving} specifically the rapidly growing biomedical research, the need for efficient and accurate information retrieval methods has become increasingly critical. The total volume of scholarly documents and the complexity of biomedical queries present substantial challenges in extracting relevant answers from vast repositories of knowledge. As the field advances, researchers are often faced with the daunting task of sifting through an extensive array of documents to obtain precise information. This highlights the necessity for robust answer extraction and document categorization methods that can enhance the accessibility of vital information. To address these challenges, this research proposes a neuro-symbolic approach that combines optimized topic modelling with advanced machine learning techniques, effectively integrating symbolic reasoning with neural representations for enhanced document retrieval and answer extraction. Specifically, we explore the efficacy of our method through three distinct configurations: the utilization of scholarly document abstract retrieval methods, golden scholarly documents abstract, and golden snippets. Our method evaluations demonstrate that our topic model-based document categorization outperforms existing methods, such as RYGH and bio-answer finder, which utilize a complex blend of techniques like BM25~\cite{DBLP:conf/bigdataconf/Ke22}, ElasticSearch~\cite{DBLP:conf/msr/KononenkoBHG14}, and various transformer models~\cite{DBLP:journals/inffus/AnwarKCS25}. This suggests that a simpler yet fine-tuned approach can lead to more effective and cost-efficient solutions for biomedical information retrieval. The primary research question in this investigation was: \textbf{How can optimized scholarly document abstract categorization and answer extraction methodologies improve the accuracy and efficiency of information retrieval in the biomedical domain?} Addressing this question is vital, as it not only enhances the precision of answer extraction but also reduces the cognitive load on researchers seeking relevant information. Furthermore, our findings reveal that even distilled smaller language models like MiniLM~\cite{DBLP:conf/nips/WangW0B0020} can effectively extract answers when fine-tuned on domain-specific data, particularly when focused on scholarly document abstracts rather than complete documents. While the comparison with the use of Large Language Models~\cite{DBLP:journals/tist/ChangWWWYZCYWWYZCYYX24} instead of the smaller MiniLM is also an interesting research avenue, it is deemed out of the scope of this research. Overall our promising results suggest a potential shift in focus for future biomedical information retrieval methods, advocating for strategies that emphasize the utility of concise scholarly abstracts. In summary, the contributions of this research comprise:
    \begin{enumerate}
        \item A novel neuro-symbolic answer extraction methodology that combines optimized topic modelling and advanced machine learning techniques, effectively addressing the challenges in biomedical information retrieval, particularly in extracting answers from an expanding corpus of scholarly documents abstract.

        \item A novel answer extraction methodology that combines optimized topic modelling and advanced machine learning techniques, effectively addressing the challenges in biomedical information retrieval, particularly in extracting answers from an expanding corpus of scholarly documents abstract.
        
        \item A comprehensive evaluation of the proposed method across three configurations— scholarly document abstract retrieval, golden scholarly documents abstract, and golden snippets—demonstrating its superior performance over existing methods like RYGH and bio-answer finder, while highlighting the advantages of simplicity and domain-specific fine-tuning.
        
        \item Insights into the effective use of distilled models, specifically MiniLM, for accurate answer extraction when fine-tuned on domain-specific data, along with recommendations for future biomedical information retrieval methods to focus on concise scholarly document abstracts for enhanced efficiency and accuracy.
    \end{enumerate}

\begin{table}[ht]
    \centering
    \begin{tabular}{p{4cm} | p{1.5cm} | p{8cm}}
        \hline
        \textbf{methods} & \textbf{Phase} & \textbf{Approach} \\
        \hline
        
        bio-answerfinder & A, B & Bio-AnswerFinder, ElasticSearch, Bio-ELECTRA, ELECTRA, BioBERT, SQuAD, wRWMD, BM25, LSTM, T5 \\
        \hline
        bioinfo & A, B & BM25, ElasticSearch, distant learning, DeepRank, universal weighting passage mechanism (UPWM), PARADE-CNN, PubMedBERT \\
        \hline
        LaRSA & A, B & ElasticSearch, BM25, SQuAD, Macro Passage Ranking, BioBERT, BoolQA, BART \\
        \hline
        ELECTROBERT & A, B & ELECTRA, ALBERT, BioELECTRA, BERT \\
        \hline
        \textbf{NEUROSYM-BIOCAT} & \textbf{A, B} & \textbf{OVB-LDA, CMA-ES, MiniLM} \\
        \hline
        RYGH & A & BM25, BioBERT, PubMedBERT, T5, BERTMeSH, SciBERT \\
        \hline
        gsl & A & BM25, BERT, dual-encoder \\
        \hline
        BioNIR & B & sBERT, distance metrics \\
        \hline
        KU-methods & B & BioBERT, data augmentation \\
        \hline
        MQ & B & tf-idf, sBERT, DistilBERT \\
        \hline
        Ir\_sys & B & BERT, SQuAD1.0, SpanBERT, XLNet, PubMedBERT, BioELECTRA, BioALBERT, BART \\
        \hline
        UDEL-LAB & B & BioM-ALBERT, Bio-ELECTRA, SQuAD \\
        \hline
        MQU & B & BART, summarization \\
        \hline
        NCU-IISR/AS-GIS & B & BioBERT, BERTScore, SQuAD, logistic-regression \\
        \hline
    \end{tabular}
    \caption{Approaches used by BioASQ 10b Participants.}
    \label{table:bioasq10b-participants}
\end{table}

\section*{Literature Review}

Latent Dirichlet Allocation (LDA)~\cite{blei2003latent} is a foundational generative probabilistic model introduced by Blei, Ng, and Jordan, designed for analyzing collections of discrete data, such as text corpora, by uncovering latent topics within documents. As a three-level hierarchical Bayesian model, LDA represents each document as a mixture over an underlying set of topics, where each topic itself is modelled as a mixture over topic probabilities, thus providing an explicit representation of the document's thematic structure. The model's primary advantage lies in its ability to capture complex intra-document statistical relationships while remaining computationally efficient through the use of variational inference techniques and an Expectation-Maximization (EM) algorithm~\cite{moon1996expectation} for parameter estimation. This approach outperforms traditional dimensionality reduction techniques like Latent Semantic Indexing (LSI)~\cite{lsi} and probabilistic LSI (pLSI)~\cite{plsi}, which lack a fully generative model for documents and suffer from issues like overfitting and limited scalability to unseen data~\cite{lda}. Building on this foundation, Hoffman et al. developed an online variational Bayes (VB) algorithm for LDA~\cite{ovblda}, optimizing its application for large-scale and streaming data scenarios. This advancement leverages online stochastic optimization to enable efficient topic modelling on massive datasets, significantly reducing computational costs compared to traditional batch algorithms while maintaining or even enhancing model accuracy. The online LDA algorithm's ability to handle document collections in a streaming fashion without the need for iterative data storage positions it as a crucial tool for real-time text analysis in various applications, making it highly relevant for contemporary machine learning and natural language processing tasks~\cite{rabby2023multi}.

The BI-population Covariance Matrix Adaptation Evolution Strategy (CMA-ES)~\cite{hansen2009benchmarking}, employs a dual restart strategy with increasing and varying small population sizes, making it an effective optimization technique for complex search landscapes. Benchmarked on the BBOB-2009 noiseless testbed, it successfully solved up to 23 out of 24 functions in different dimensions within a constrained budget of evaluations~\cite{cmaes}. While demonstrating strong performance on unimodal and multimodal functions, its limitations on smooth, regular, and separable functions highlight areas for improvement, suggesting that refining its restart schedule could further enhance its optimization capabilities. In Task 10b of the BioASQ challenge, various approaches were employed in biomedical question answering and document categorization. A total of 70 methods were submitted across different research groups, showcasing diverse techniques in solving the tasks~\cite{bioasq_overview}. The challenge was divided into two phases: Phase A, which focused on information retrieval, and Phase B, which addressed question answering. Below is an overview of the key methodologies and advancements in these fields. The method developed by UCSD, \textit{bio-answerfinder}, evolved from earlier work, introducing improvements in tokenization and query expansion~\cite{zyurt2021EndtoendBQ}. In Phase A, UCSD employed a relaxation mechanism in the ranked keyword-based document retrieval process, using BM25 retrieval and ranking keywords with a cascade of LSTM layers~\cite{bioasq_overview}~\cite{lstm}. In Phase B, they shifted towards abstractive summarization using T5, marking a transition from extractive to generative approaches in summarization tasks~\cite{bioasq_overview}. The University of Aveiro leveraged their transformer-UPWM model~\cite{AlmeidaMatos_CLEF_2021}, extending it with the PARADE-CNN architecture~\cite{DBLP:journals/corr/abs-2008-09093}. Throughout both phases, they utilized the PubMedBERT transformer model, demonstrating the effectiveness of transformers in biomedical text representation. In Phase B, they introduced a classifier for handling yes/no questions, indicating an efficient approach to binary classification~\cite{bioasq_overview}. Mohamed I University’s \textit{LaRSA} method combined ElasticSearch with BM25 for retrieval, alongside fine-tuning Roberta-base and BioBERT models for various question types~\cite{bioasq_overview}. Their use of cross-encoders and re-rankers, trained on the MS Marco Passage Ranking task, enhanced their retrieval accuracy, while BART models were employed for generating ideal answers in Phase B, demonstrating an integrated approach to question answering~\cite{bioasq_overview}. The \textit{ELECTROBERT} method, designed by BSRC Alexander Fleming, merged the strengths of ELECTRA and ALBERT, focusing on document retrieval tasks. Pre-training on PubMed abstracts and fine-tuning on the BioASQ9 dataset emphasized the relevance of combining large-scale transformer models with domain-specific data for better document ranking~\cite{bioasq9}. \textit{RYGH} used a multi-stage retrieval method combining BM25 with BioBERT, PubMedBERT, and SciBERT, integrating both traditional and modern retrieval techniques to enhance biomedical question answering~\cite{bioasq_overview}. This hybrid approach highlights the importance of combining transformer models with established retrieval algorithms. Google’s \textit{gsl} methods applied a zero-shot hybrid model that integrated retrieval and re-ranking stages. Their approach combined BM25 for document retrieval with a dual-encoder and cross-attention re-ranking model, reflecting the power of multi-stage architectures in optimizing biomedical document retrieval~\cite{bioasq_overview}. TU Kaiserslautern introduced the \textit{BioNIR} method, based on sentence-BERT (sBERT), which encoded queries and abstracts sentence-by-sentence and used distance metrics for retrieval, underscoring the utility of sentence-level ranking in biomedical text processing~\cite{bioasq_overview}. KU methods focused on augmenting their training data in Phase B using a BioBERT backbone. They applied question-generation techniques to increase the diversity of their training dataset, exemplifying the use of data augmentation in improving model generalization for question-answering tasks~\cite{bioasq_overview}. Macquarie University’s two teams used different models for different tasks. The \textit{MQ} method used DistilBERT for ideal answers, while \textit{MQU} employed BART-based abstractive summarization methods, showcasing a dual strategy of lightweight and abstractive models to address various question types effectively~\cite{bioasq_overview}. Fudan University’s \textit{Ir\_sys} methods employed a combination of BERT, BioBERT, and BART models across multiple question-answering tasks. This comprehensive approach across factoid, list, and summarization tasks indicates the flexibility of transformer models in biomedical question answering~\cite{bioasq_overview}. The University of Delaware’s \textit{UDEL-LAB} method focused on fine-tuning transformer-based models like BioM-ALBERT and BioM-ELECTRA. Their emphasis on hyperparameter tuning and analysis of randomness in transformers illustrates their deep focus on model optimization for biomedical QA~\cite{bioasq_overview}. The \textit{NCU-IISR/AS-GIS} team used BioBERT with logistic regression to rank snippets and generate answers. Their move from traditional metrics like ROUGE-SU4 to BERTScore indicates a shift toward context-aware evaluation in summarization tasks~\cite{bioasq_overview}. Finally, the \textit{OAQA} baseline method, based on traditional NLP techniques like MetaMap and TmTool, provided a point of comparison for modern deep learning models. This method continues to serve as a useful benchmark in BioASQ, ensuring that innovations are evaluated against established methods~\cite{bioasq_overview}. Table~\ref{table:bioasq10b-participants} provides an overview of the approaches utilized by participants in the BioASQ 10b competition.

        \begin{figure}[ht]
            \centering
            \includegraphics[height=12cm, width=1.0\textwidth]{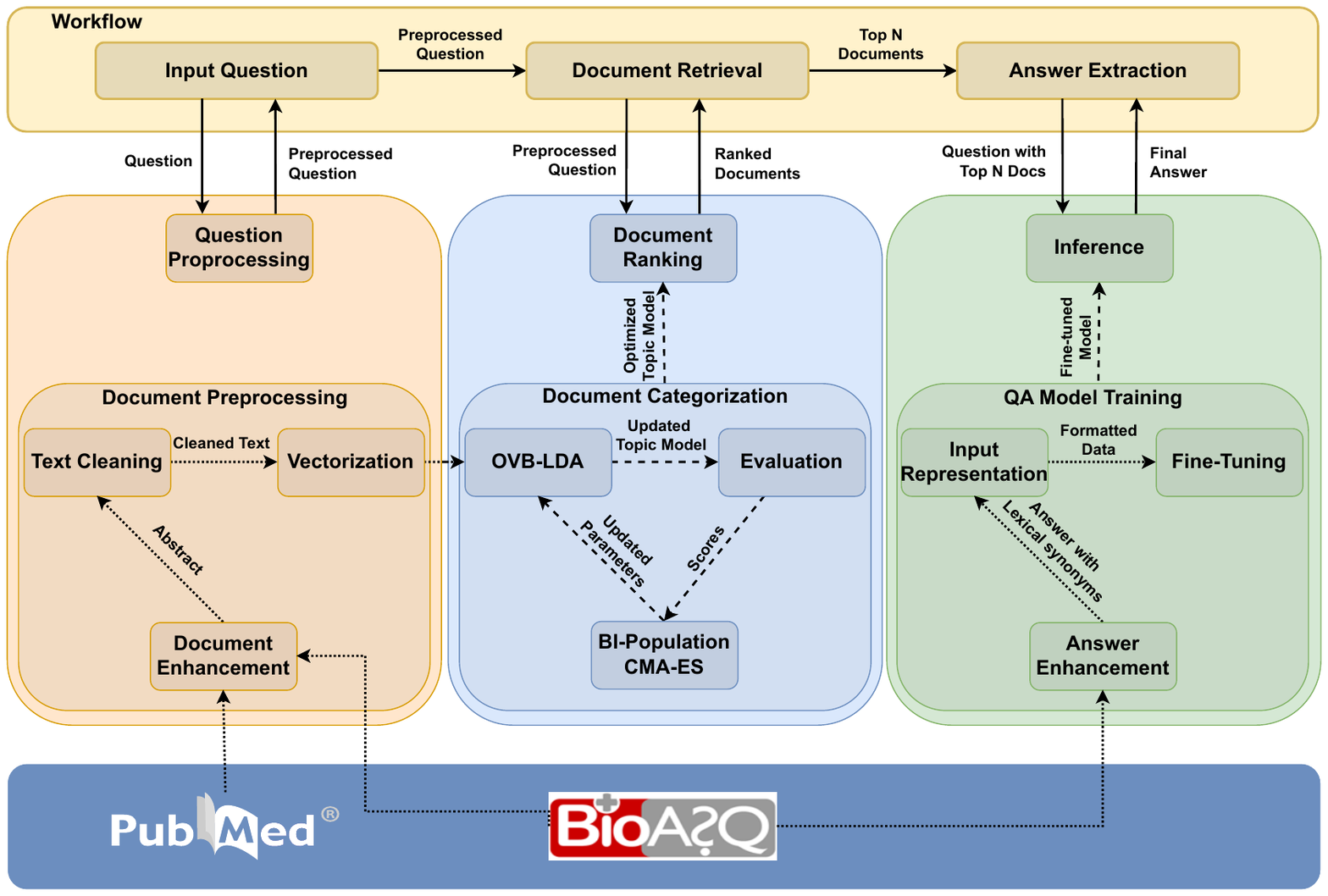}
            \caption{Overview of Methodology.}
            \label{fig:methodology-overview}
        \end{figure}

\section*{Methods}
  
    Our research employed two distinct approaches to extract answers for factoid and list-type questions from biomedical scholarly documents abstracts. The methodology consisted of two different phases: scholarly document abstract categorization and answer extraction. Subsequent sections provide comprehensive explanations and detailed procedures for each phase. Figure~\ref{fig:methodology-overview} illustrates an overview of the overall methodology.

        \begin{figure}[ht]
            \centering
            \includegraphics[height=8.5cm, width=1.0\textwidth]{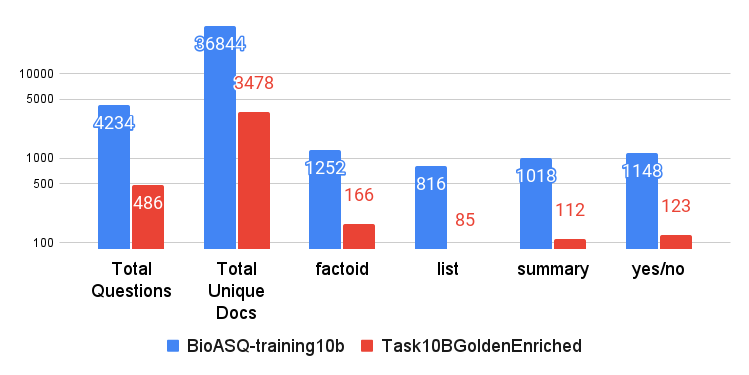}
            \caption{{Overview of BioASQ10 Datasets.}}
            \label{fig:dataset-overview}
        \end{figure}

    \subsection*{Corpora Details}
            
       In this experiment, we employed the BioASQ10 dataset~\cite{bioasq_overview} as our primary data source. The dataset is divided into two sections: the "BioASQ-training10b" training dataset and the "Task10BGoldenEnriched" test dataset. The "BioASQ-training10b" dataset is comprehensive, containing 4,234 questions along with an extensive collection of 36,844 unique biomedical scholarly document abstracts. These questions are categorized into four types: factoid, list, summary, and binary. However, our experiment focused exclusively on factoid (1,252) and list-type (816) questions, as these are particularly aligned with our research objectives. The "Task10BGoldenEnriched" dataset, used for evaluation, consists of 486 questions and a corresponding set of 3,478 unique scholarly document abstracts. From this dataset, we specifically utilized factoid (166) and list-type (85) questions, reflecting the question types in the training dataset. This selection ensured that our evaluation was conducted on comparable question categories.        
       Figure~\ref{fig:dataset-overview} provides an overview of the overall dataset.

    \subsection*{Data Preprocessing} 
    
        We begin by enhancing the dataset by fetching PubMed~\cite{DBLP:journals/frma/Lezhnina23} abstracts corresponding to the PubMed IDs provided in the BioASQ10 dataset. Following this, we conducted several essential preprocessing steps to ensure the quality and consistency of the data for our experiments. We removed stop words—commonly occurring words in natural language that carry minimal semantic value~\cite{DBLP:journals/jiis/RabbyB23}. This step effectively reduced noise in the dataset, improving the efficiency of subsequent analyses. In addition to stopping word removal, we applied a comprehensive text-cleaning process, which included eliminating punctuation and special characters. By removing these extraneous symbols, we created a more coherent and standardized dataset, ensuring uniformity across the data. Finally, we applied a stemming process to reduce words to their root forms, which allowed for more accurate and meaningful analyses by grouping related terms under a common base form. These preprocessing steps collectively contributed to the creation of a cleaner, more structured dataset that was better suited for the analytical tasks in our experiment.

    \subsection*{Document Categorization}

        To achieve robust results for document categorization, we employed an advanced approach that integrates state-of-the-art techniques and optimization strategies. For the categorization of scholarly documents abstract, we applied the Online Variational Bayes for Latent Dirichlet Allocation (OVB-LDA)~\cite{ovblda}, a topic modelling method well-suited for identifying latent topics within large corpora. OVB-LDA's ability to uncover hidden structures in the dataset allowed us to categorize documents based on their underlying topic effectively. To further enhance the performance of OVB-LDA, we employed the Bimodal Population Covariance Matrix Adaptation Evolution Strategy (BI-POP CMA-ES)~\cite{hansen2009benchmarking}, a cutting-edge optimization method known for effectiveness in exploring high-dimensional parameter spaces. This strategy was critical in fine-tuning OVB-LDA, significantly improving both the accuracy and efficiency of the document categorization process. The combination of these advanced techniques allowed us to build a highly accurate model for the retrieval of scholarly document abstracts.

        \subsubsection*{Feature Engineering} 

        For feature engineering, we utilized the Bag-of-Words (BoW)~\cite{DBLP:journals/mlc/ZhangJZ10} method, a well-established and widely-used technique in natural language processing. The BoW approach generates a representation of each document by capturing the occurrence and frequency of individual words within the scholarly document's abstract. This transformation created high-dimensional vectors, with each vector encapsulating the unique linguistic characteristics and content nuances of its respective document. These feature vectors formed the primary input for our OVB-LDA method, facilitating the categorization of scholarly document abstracts based on their latent topics. By converting unstructured abstracts into a structured and quantifiable format, BoW played a crucial role in empowering OVB-LDA to effectively process and categorize scholarly document abstracts. This feature engineering step was essential in linking our raw data with the advanced categorization techniques, ensuring that the data was in a suitable form for robust topic modelling and subsequent analysis.

        \begin{algorithm}[H]
            \caption{Optimizing  OVB-LDA leveraged Document Categorization with BI-POP CMA-ES }
            \label{algo:document_retrieval}
            \begin{algorithmic}[1]
                \State \textbf{Initialization}:
                \State \hspace{\algorithmicindent} - Initialize OVB-LDA model parameters: $\alpha$, $\beta$.
                \State \hspace{\algorithmicindent} - Initialize BI-POP CMA-ES parameters: $\lambda$ (population size), $\mu$ (number of selected solutions), $c$ (covariance matrix), $\sigma$ (step size), $p_{\text{step}}$ (mutation probability).
                \State \hspace{\algorithmicindent} - Set maximum iterations $T$ and initial population size $N$.
                
                \State \textbf{Preprocessing}:
                \State \hspace{\algorithmicindent} - Tokenize and preprocess the document corpus: $D = \{d_1, d_2, \dots, d_n\}$.
                \State \hspace{\algorithmicindent} - Construct a dictionary $V$ of unique tokens from the corpus.
                \State \hspace{\algorithmicindent} - Represent each document $d_i$ as a Bag-of-Words (BoW) vector $v_i \in \mathbb{R}^{|V|}$.
                
                \State \textbf{Optimization with BI-POP CMA-ES and Document Retrieval using OVB-LDA}:
                \State \hspace{\algorithmicindent} - Initialize the population of solutions $\{\theta_1, \theta_2, \dots, \theta_\lambda\}$ where $\theta_i$ represents OVB-LDA parameters.
                \State \hspace{\algorithmicindent} - Evaluate the fitness $f(\theta_i)$ of each solution $\theta_i$ using F1 score.
                \State \hspace{\algorithmicindent} - Sort the population based on fitness.
                \State \hspace{\algorithmicindent} - \textbf{For} $t = 1$ to $T$ iterations:
                \State \hspace{\algorithmicindent} \hspace{\algorithmicindent} - Select the top $\mu$ solutions: $\{\theta_1, \theta_2, \dots, \theta_\mu\}$.
                \State \hspace{\algorithmicindent} \hspace{\algorithmicindent} - Recombine and mutate the selected solutions to generate offspring.
                \State \hspace{\algorithmicindent} \hspace{\algorithmicindent} - Evaluate the fitness of offspring solutions: $f(\theta_{\text{offspring}})$.
                \State \hspace{\algorithmicindent} \hspace{\algorithmicindent} - Select the top $\lambda$ solutions from the combined pool of parents and offspring.
                \State \hspace{\algorithmicindent} - Update BI-POP CMA-ES parameters: $c$, $\sigma$, $p_{\text{step}}$.
                
                %\State \textbf{Document Retrieval using OVB-LDA}:
                \State \hspace{\algorithmicindent} - Train OVB-LDA model on the document corpus $D$ using the optimized parameters $\hat{\alpha}$, $\hat{\beta}$.
                \State \hspace{\algorithmicindent} - For a given query $q$, transform it into a topic distribution $\theta_q$ using the trained OVB-LDA model.
                \State \hspace{\algorithmicindent} - Compute cosine similarity: 
                \[
                \text{sim}(q, d_i) = \frac{\theta_q \cdot \theta_{d_i}}{\|\theta_q\| \|\theta_{d_i}\|}
                \]
                between the query's topic distribution $\theta_q$ and each document's topic distribution $\theta_{d_i}$.
                \State \hspace{\algorithmicindent} - Rank documents $d_i$ based on similarity scores $\text{sim}(q, d_i)$.
                
                \State \textbf{Output}:
                \State \hspace{\algorithmicindent} - Return ranked list of documents $[d_1, d_2, \dots, d_k]$ based on relevance to the query.
            \end{algorithmic}
        \end{algorithm}
        
        \subsubsection*{Document Categorization using OVB-LDA}
        
        Building upon the feature engineering step, where we transformed scholarly document abstracts into structured vectors using the BoW method, our research employed the Online Variational Bayes for Latent Dirichlet Allocation (OVB-LDA)~\cite{hoffman2010online} as a robust and data-efficient method for scholarly documents abstract categorization and information retrieval. OVB-LDA's strength lies in its ability to efficiently handle large-scale collections of scholarly documents abstract by processing data in manageable mini-batches, making it well-suited for our extensive dataset. OVB-LDA extracting scholarly documents abstract-topic distributions from the scholarly documents abstracts. Each scholarly document abstract in our corpus was associated with a unique topic distribution, reflecting its underlying thematic structure. Simultaneously, queries were transformed into topic distributions using the same LDA approach, allowing for a consistent representation of both documents and queries. For the scholarly documents abstract retrieval process, we calculated the cosine similarity between the topic distribution of a query and those of the scholarly document abstracts. This method leverages the inherent thematic structure captured by the topic distributions to retrieve the most relevant documents efficiently and accurately. OVB-LDA's ability to manage large-scale document collections providing precise categorization and retrieval emphasizes its significant role in our research~\cite{DBLP:journals/tkde/ZengLC16}. 

        \begin{table}[htbp]
            \centering
            \begin{tabular}{cccc}
                \cline{1-4}
                \textbf{Hyperparameter} & \textbf{Input Search Space} & \textbf{Optimal Value} & \textbf{Importance} \\
                \cline{1-4}
                num\_topics & 20-2000 & 1241 & 0.49 \\
                chunksize & 1-4096 & 2877 & 0.47 \\
                passes & 1-10 & 5 & 0.01 \\
                decay & 0.5-1.0 & 0.5 & 0.01 \\
                eval\_every & 1-10 & 10 & 0.01 \\
                iterations & 5-200 & 188 & 0.01 \\
                \cline{1-4}
            \end{tabular}
            \caption{Input Search Space, Optimal Parameters, and Hyperparameter Importance.}
            \label{table:hyperparameter}
        \end{table}

        \subsubsection*{Optimization Method for Scholarly Document Categorization}
        
        Following the implementation of the OVB-LDA method for scholarly document abstract categorization, we introduced an advanced optimization approach to further enhance the performance of our scholarly document abstract retrieval method. Specifically, we utilized the Bimodal Population Covariance Matrix Adaptation Evolution Strategy (BI-POP CMA-ES)~\cite{he2018enhancing}, a cutting-edge optimization technique designed to fine-tune the parameters of OVB-LDA for improved categorization accuracy. The primary goal of applying BI-POP CMA-ES in our experiment was to maximize the retrieval performance of OVB-LDA by iteratively adjusting its parameters. This process involved evaluating the performance of various parameter configurations across a diverse set of queries, allowing for precise calibration based on the specific characteristics of both the dataset and the query corpus. By continuously refining these parameters, BI-POP CMA-ES enabled OVB-LDA to perform more effectively in retrieving relevant scholarly document abstracts. One of the unique advantages of BI-POP CMA-ES is its dynamic adaptation of population size during the optimization process~\cite{DBLP:conf/gecco/LoshchilovSS13a}. This feature allowed the method to navigate the complex optimization landscape efficiently, making it particularly suited for the intricate task of biomedical scholarly document abstract categorization. Through the application of BI-POP CMA-ES, we achieved optimal parameter configurations for OVB-LDA, ultimately enhancing both the accuracy and efficiency of our scholarly document abstract categorization and retrieval method. The final optimized parameters are given in Table~\ref{table:hyperparameter}. The detailed steps of this process are methodically presented in Algorithm~\ref{algo:document_retrieval}.

\subsection*{Answer Extraction}

The answer extraction phase in our method focuses on retrieving precise answers from the categorized biomedical scholarly documents abstract, building upon the scholarly document abstract categorization process discussed in previous sections. For this task, we employed the Transformer architecture, which is powered by the self-attention mechanism~\cite{attention}. This architecture enabled the efficient extraction of relevant information from the large-scale biomedical scholarly document corpus. To enhance efficiency and maintain performance, we integrated the MiniLM model~\cite{minilm}, a distilled version of the larger Transformer model, which emphasizes the use of deep self-attention layers while maintaining computational efficiency. To adapt the model specifically for answering biomedical factoid and list-type questions, we fine-tuned the MiniLM model on a domain-specific dataset of biomedical scholarly document abstracts. This fine-tuning step allowed the model to understand better the specialized language and entities found in biomedical literature, ensuring more accurate and relevant answer extraction~\cite{DBLP:journals/bmcbi/SuV22}.

    \subsubsection*{Lexical Synonyms Extraction}
    \label{convert-to-squad}

        In handling factoid and list-type queries, we further enriched the answer extraction process by incorporating lexical synonym extraction. This process was vital for enhancing the model's ability to identify different variations of biomedical terms and entities. For this purpose, we utilized the WordNet lexical database~\cite{wordnet} ~\cite{DBLP:journals/jifs/JumdeK24}, which is widely used for identifying synonyms across various linguistic contexts. By applying WordNet to the biomedical dataset, we methodically expanded the synonym coverage for key biomedical entities found in the scholarly document abstracts~\cite{DBLP:conf/ihi/LiuMPMP12}. This enriched the dataset with semantically related terms and improved the method's capacity to match queries with relevant answers, even when different terminology or synonyms were used. This synonym extraction approach was applied to both factoid and list-type queries from our dataset, significantly improving the quality and comprehensiveness of the data.

        \subsubsection*{Input Representation for Fine-tuning}

        In our research, we employed the Transformer architecture, specifically leveraging the MiniLM model and its variant, "$\text{minilm-uncased-squad2}$"~\cite{d2023evaluating}. This model was used for the tokenization and was essential for the effective fine-tuning of our answer extraction process. To maintain a robust connection between the tokenized data and the original scholarly documents abstract, we established a $sample\_mapping$ structure. This mapping allowed us to trace back each token to its corresponding document, ensuring that the context of the extracted answers was preserved. To facilitate the alignment of tokens with their respective character positions within the scholarly document abstracts, we utilized offset mapping. This approach allowed us to accurately represent the position of each token in the original text, which is crucial for determining the start and end positions of the answers extracted from the scholarly documents abstract. Furthermore, we ensured that each tokenized scholarly document abstract was labelled with start and end positions that delineate the exact span of the answer within the text. For instances where no answer was present, we labelled these cases using the CLS token index. This methodatic labelling process was vital for training the MiniLM model, enabling it to learn effectively from both positive (answer-present) and negative (answer-absent) examples. By establishing this structured input representation, we connected the answer extraction phase with the previous stages of scholarly document abstract categorization and lexical synonym extraction, ultimately enhancing the model's performance in retrieving accurate and relevant answers from biomedical literature.

       \begin{algorithm}[htbp]
            \caption{Fine-tuning}
            \label{algo:fine-tuning}
            \begin{algorithmic}[1]
            \State \textbf{Input:} Tokenized context \( c \), question \( q \), scholarly document abstract \( \mathcal{D} \), and a pre-trained model \( \mathcal{M} \)
            \State \textbf{Initialization}:
            \State \hspace{\algorithmicindent} Load the pre-trained model (\( \mathcal{M} \))
            \State \hspace{\algorithmicindent} Define the optimization algorithm (e.g., Adam) with appropriate hyperparameters
            
            \State \textbf{For} epoch = 1 to \( \mathcal{E} \) \textbf{do:}
            \State \hspace{\algorithmicindent} \textbf{For each} example \( (c, q) \) in \( \mathcal{D} \) \textbf{do:}
            \State \hspace{\algorithmicindent} \hspace{\algorithmicindent} \textbf{Forward Pass:}
            \State \hspace{\algorithmicindent} \hspace{\algorithmicindent} \( a = \mathcal{M}(c, q) \)
            \State \hspace{\algorithmicindent} \hspace{\algorithmicindent} \textbf{Loss Calculation:} 
            \State \hspace{\algorithmicindent} \hspace{\algorithmicindent} \( \mathcal{L} = \text{CrossEntropyLoss}(a) \)
            \State \hspace{\algorithmicindent} \hspace{\algorithmicindent} \textbf{Backward Pass:} 
            \State \hspace{\algorithmicindent} \hspace{\algorithmicindent} Compute the gradients: \( \nabla \mathcal{L} \)
            \State \hspace{\algorithmicindent} \hspace{\algorithmicindent} \textbf{Parameter Update:}
            \State \hspace{\algorithmicindent} \hspace{\algorithmicindent} Update the model's weights: \( \mathcal{M} \leftarrow \mathcal{M} - \alpha \nabla \mathcal{L} \) 
            \State \hspace{\algorithmicindent} \textbf{End For}
            \State \textbf{Return:} the fine-tuned model \( \mathcal{M}' \)
            \end{algorithmic}
        \end{algorithm}

    \subsubsection*{Fine-tuning}

        For the QA task, our primary objective was to predict the start and end positions of the answer span within scholarly document abstracts. In this context, let \( \mathcal{D} \) denote a scholarly document abstract, \( \mathcal{M} \) represent the pre-trained model, \( c \) be the tokenized context, and \( q \) signify the query question. The predicted span \( a \) is computed as:

        \[
        a = \mathcal{M}(c, q)
        \]
        
        To evaluate the performance of the model, we calculate the loss \( \mathcal{L} \) using the cross-entropy loss function:
        
        \[
        \mathcal{L} = \text{CrossEntropyLoss}(a)
        \]
        
        % The gradients \( \nabla \mathcal{L} \) are then computed as:
        
        % \[
        % \nabla \mathcal{L}
        % \]
        
        The learning rate is denoted by \( \alpha \) and gradients are denoted by \( \nabla \mathcal{L} \) , the updated model weights are obtained through the following update rule:
        
        \[
        \mathcal{M} \leftarrow \mathcal{M} - \alpha \nabla \mathcal{L}
        \]
        
        The fine-tuning process aims to predict two distinct probability distributions over positions in the scholarly document abstract: one for the start position and another for the end position of the answer span. This dual prediction framework enables the model to effectively identify the boundaries of the answer within the text. Algorithm~\ref{algo:fine-tuning} outlines the fine-tuning process for the QA task, integrating seamlessly with our previous sections on input representation and the use of the MiniLM model. By refining the model using this structured approach, we enhance its ability to accurately retrieve answers from the categorized scholarly documents abstract.

 \begin{algorithm}[H]
    \caption{Answer Extraction}
    \label{algo:answer-extraction}
    \begin{algorithmic}[1]
        \State \textbf{Preprocessing}:
        \State \hspace{\algorithmicindent}For each paragraph \( p \) in \( D \):
        \State \hspace{\algorithmicindent} \quad \( \text{Basket}(p, Q) \leftarrow \text{CreateBasket}(p, Q) \)
        \State \hspace{\algorithmicindent}For each \( \text{Basket} \):
        \If{$\text{length}(p) > L$}
            \State \hspace{\algorithmicindent} \quad \( \text{Sample}(p_i, Q) \leftarrow \text{SplitIntoSamples}(p, Q, L) \)
            \State \hspace{\algorithmicindent} \quad Tokenize \( Q \) and \( p_i \) to get tokens \( T_Q \) and \( T_{p_i} \)
        \EndIf
        \State \textbf{Modeling}:
        \State \hspace{\algorithmicindent}For each \( \text{Sample} \):
        \State \hspace{\algorithmicindent} \quad \( V \leftarrow \mathcal{M'}(T_Q, T_{p_i}) \) where \( V \) is the contextualized word vector
        \State \hspace{\algorithmicindent} \quad Generate logits:
        \State \hspace{\algorithmicindent} \quad \( \text{Logits}_{\text{start}}, \text{Logits}_{\text{end}} \leftarrow \text{FeedForward}(V) \)
        
        \State \textbf{Aggregation}:
        \State \hspace{\algorithmicindent}For each \( \text{Sample} \):
        \If{$\text{Logits}_{\text{start}}[0]$ and $\text{Logits}_{\text{end}}[0]$ are maximum}
            \State \hspace{\algorithmicindent} \quad \( A \leftarrow \text{"No-answer"} \)
        \Else
            \State \hspace{\algorithmicindent} \quad \( A \leftarrow \text{ExtractAnswer}(\text{Logits}_{\text{start}}, \text{Logits}_{\text{end}}, T_{p_i}) \)
        \EndIf
        \State \textbf{Formatting Predictions}:
        \State \hspace{\algorithmicindent}Extract string prediction:
        \State \hspace{\algorithmicindent} \quad \( A_{\text{string}} \leftarrow \text{TokenToString}(A) \)
        \State \hspace{\algorithmicindent}Determine character index:
        \State \hspace{\algorithmicindent} \quad \( \text{Index} \leftarrow \text{FindStartIndex}(A_{\text{string}}, D) \)
        
        \State \textbf{Return}: \( A_{\text{string}} \) and \( \text{Index} \)
    \end{algorithmic}
\end{algorithm}

    \subsubsection*{Answers Extraction}

        To extract relevant answers from the categorized scholarly documents abstract, we utilized a specialized reader from the Haystack library~\cite{huynh2002haystack}. This extraction process begins by fragmenting each scholarly document abstract into manageable samples, especially if the scholarly documents abstract exceed the model's maximum sequence length. Each sample, along with the corresponding question, is then tokenized and passed through the fine-tuned model to obtain word vectors. These word vectors are crucial as they are transformed into logits, which indicate potential answer spans within the scholarly document abstract. Subsequently, we categorize these logits to determine the most likely answer span or conclude that no answer exists within the scholarly document abstract. This methodatic approach to answer extraction builds on the previous sections detailing the model fine-tuning and input representation processes. By leveraging the capabilities of the Haystack library and the optimized model, we aim to achieve precise and relevant answer retrieval from biomedical scholarly document abstracts, thereby enhancing the overall effectiveness of our QA method. The detailed steps of this process are methodically presented in Algorithm~\ref{algo:answer-extraction}.

   \subsection*{Evaluation Metrics}

   We also explores the evaluation metrics used for Scholarly document abstract categorization and answer extraction, particularly for list-type and factoid questions. Precision, recall, and F1-score are employed to assess categorization and entity identification accuracy in list questions, while strict and lenient accuracy is applied to factoid questions~\cite{DBLP:conf/eval4nlp/YacoubyA20} ~\cite{DBLP:journals/tois/Lin07}. Additionally, the Mean Reciprocal Rank (MRR) serves as the official metric for ranking the correctness and relevance of factoid answers.

        \subsubsection*{Scholarly Document Categorization and List Questions}
        
        In the context of Scholarly document abstract categorization and answer extraction for list-type questions, we utilized the following computation metrics:
        
        \begin{itemize}
            \item \textbf{Precision (P)}: For Scholarly document abstract categorization, precision measures the proportion of correctly identified scholarly document abstracts, and for list-type questions, entities in the method's answer list relative to the total number of entity names in the golden list. It can be quantified as:
            
            \[
            P = \frac{TP}{TP + FP}
            \]
            
            Here, \(TP\) represents the count of true positives, and \(FP\) represents the count of false positives.
            
            \item \textbf{Recall (R)}: Recall measures the proportion of correctly identified Scholarly document abstracts in the case of retrieval and entity names in the case of list questions in the method's response relative to golden Scholarly document abstracts and lists. It is defined as:
            
            \[
            R = \frac{TP}{TP + FN}
            \]
            
            Where \(TP\) represents the count of true positives, and \(FN\) represents the count of false negatives.
            
            \item \textbf{F-measure (F1)}: The F1-score is the harmonic mean of precision and recall, offering a balanced measure of both metrics. It can be calculated as:
            
            \[
            F1 = 2 \times \frac{P \times R}{P + R}
            \]
        \end{itemize}
        
        \subsubsection*{Factoid Questions}
        
        In the context of factoid questions, methods are tasked with returning up to five entity names, ranked by confidence. To assess the performance of these methods, the BioASQ team provides a golden entity name for each question along with its potential synonyms. Evaluation involves the use of two key accuracy metrics:
        
        \begin{itemize}
            \item \textbf{Strict Accuracy (SAcc)}: A question is considered correctly answered if the golden entity name or one of its synonyms is the topmost element in the list of returned entities. It can be quantified as:
            
            \[
            SAcc = \frac{c1}{n}
            \]
            
            Where \(c1\) represents the count of questions correctly answered with strict accuracy, and \(n\) is the total number of questions.
            
            \item \textbf{Lenient Accuracy (LAcc)}: This metric allows for a more flexible evaluation criterion. A question is considered correctly answered if the golden entity name or one of its synonyms appears anywhere in the list of returned entities. Its calculation is given by:
            
            \[
            LAcc = \frac{c5}{n}
            \]
            
            Here, \(c5\) represents the count of questions correctly answered with lenient accuracy, and \(n\) is again the total number of questions.
        \end{itemize}
        
        For factoid questions, the official metric used for evaluation is the \textbf{Mean Reciprocal Rank (MRR)}. It is computed as follows:
        
        \[
        MRR = \frac{1}{n} \sum_{i=1}^{n} \frac{1}{r(i)}
        \]
        
        Where \(n\) is the total number of questions, and \(r(i)\) is the rank of the first correct answer for the \(i\)-th question in the list of returned entities. MRR provides a comprehensive method performance measure, considering the accuracy and rank of correct answers.

        \begin{longtable}{>{\raggedright\arraybackslash}p{0.35\textwidth} 
                            >{\centering\arraybackslash}p{0.15\textwidth} 
                            >{\centering\arraybackslash}p{0.15\textwidth} 
                            >{\centering\arraybackslash}p{0.15\textwidth}}
           
            \toprule
            \thead{method} & \thead{Mean Precision} & \thead{Recall} & \thead{F-Measure} \\
            \midrule
            \endfirsthead
        
            \multicolumn{4}{c}{{\tablename\ \thetable{} Score Comparison of Document Categorization (continued)}} \\
            \toprule
            \thead{method} & \thead{Mean Precision} & \thead{Recall} & \thead{F-Measure} \\
            \midrule
            \endhead
        
            \bottomrule
            \endfoot
            
            \bottomrule
            \endlastfoot
        
            \textbf{Batch 1} & & & \\
            bio-answerfinder & 0.39 & 0.41 & 0.35 \\
            \textbf{Proposed} & \textbf{0.30} & \textbf{0.54} & \textbf{0.31} \\
            RYGH-1 & 0.28 & 0.61 & 0.29 \\
            RYGH-4 & 0.27 & 0.61 & 0.29 \\
            RYGH-3 & 0.27 & 0.61 & 0.29 \\
            RYGH & 0.27 & 0.59 & 0.29 \\
            The basic end-to-end & 0.24 & 0.51 & 0.27 \\
            Basic e2e mid speed & 0.23 & 0.49 & 0.26 \\
            \hline
        
            \textbf{Batch 2} & & & \\
            bio-answerfinder & 0.34 & 0.40 & 0.30 \\
            \textbf{Proposed} & \textbf{0.29} & \textbf{0.52} & \textbf{0.28} \\
            RYGH-3 & 0.28 & 0.57 & 0.28 \\
            RYGH-4 & 0.28 & 0.56 & 0.27 \\
            RYGH & 0.28 & 0.56 & 0.27 \\
            RYGH-1 & 0.27 & 0.58 & 0.27 \\
            bio-answerfinder-2 & 0.28 & 0.41 & 0.25 \\
            bio-answerfinder-4 & 0.27 & 0.39 & 0.24 \\
            \hline
        
            \textbf{Batch 3} & & & \\
            bio-answerfinder & 0.37 & 0.48 & 0.34 \\
            The basic end-to-end & 0.30 & 0.51 & 0.31 \\
            BioNIR Prepro-mid & 0.30 & 0.50 & 0.30 \\
            Basic e2e mid speed & 0.30 & 0.51 & 0.30 \\
            bio-answerfinder-3 & 0.33 & 0.43 & 0.30 \\
            \textbf{Proposed} & \textbf{0.29} & \textbf{0.57} & \textbf{0.30} \\
            RYGH-5 & 0.25 & 0.65 & 0.28 \\
            RYGH & 0.26 & 0.64 & 0.28 \\
            \hline
        
            \textbf{Batch 4} & & & \\
            bio-answerfinder & 0.39 & 0.43 & 0.33 \\
            \textbf{Proposed} & \textbf{0.30} & \textbf{0.53} & \textbf{0.30} \\
            bio-answerfinder-4 & 0.29 & 0.49 & 0.28 \\
            Basic e2e mid speed & 0.30 & 0.45 & 0.28 \\
            RYGH-4 & 0.27 & 0.59 & 0.28 \\
            RYGH-3 & 0.27 & 0.60 & 0.28 \\
            RYGH-1 & 0.26 & 0.60 & 0.28 \\
            bio-answerfinder-2 & 0.28 & 0.48 & 0.28 \\
            \hline
        
            \textbf{Batch 5} & & & \\
            bio-answerfinder & 0.45 & 0.43 & 0.38 \\
            \textbf{Proposed} & \textbf{0.33} & \textbf{0.47} & \textbf{0.31} \\
            bio-answerfinder-3 & 0.34 & 0.34 & 0.29 \\
            RYGH & 0.28 & 0.55 & 0.28 \\
            RYGH-4 & 0.28 & 0.55 & 0.28\\
            bio-answerfinder-2 & 0.30 & 0.40 & 0.28 \\
            RYGH-3 & 0.27 & 0.54 & 0.27 \\
            RYGH-1 & 0.27 & 0.54 & 0.27 \\
            \hline
        
            \textbf{Batch 6} & & & \\
            \textbf{Proposed} & \textbf{0.55} & \textbf{0.42} & \textbf{0.43} \\
            gsl\_zs\_rrf3 & 0.41 & 0.33 & 0.32 \\
            gsl\_zs\_nn & 0.41 & 0.32 & 0.31 \\
            RYGH-4 & 0.40 & 0.32 & 0.31 \\
            RYGH & 0.39 & 0.32 & 0.31 \\
            RYGH-1 & 0.37 & 0.30 & 0.30 \\
            gsl\_zs\_rrf1 & 0.38 & 0.30 & 0.30 \\
            gsl\_zs\_rrf2 & 0.37 & 0.30 & 0.29 \\
            \toprule
            
          \caption{Score Comparison of Document Categorization.} 
          \label{tab:DR-score-comparison} 
        \end{longtable}

\section*{Results and Discussion}

In this section, we present a comprehensive evaluation of our proposed method for answer extraction and scholarly document abstract categorization within the biomedical domain. We assess the performance across three distinct configurations: scholarly document abstract retrieval, golden scholarly documents abstract, and golden snippets. The evaluation is conducted over multiple batches of scholarly document abstracts, and the results highlight significant improvements in precision, recall, and F-measure across various question types. Notably, our method demonstrates a strong ability to retrieve accurate information, with golden snippets consistently achieving the highest performance, particularly in terms of MRR. These findings emphasize the versatility and robustness of the proposed approach across diverse datasets and evaluation metrics.

    \subsection*{Predictive Performance}

    The proposed method exhibits strong predictive performance, achieving high precision and recall across multiple batches, particularly in scholarly document abstract retrieval and golden scholarly document abstract configurations. For answer extraction, the method excels when leveraging golden snippets, yielding superior MRR and F-measure scores, especially in factoid and list-type questions. This highlights the method's capability to retrieve and extract accurate biomedical information effectively across diverse query types.

    \subsubsection*{Document Retrieval}

        In a comprehensive evaluation of scholarly document abstract categorization conducted across six different batches, the assessment was grounded on three distinct metrics: mean precision, recall, and the F-measure, which amalgamates both precision and recall to provide a holistic view of performance (as shown in Table \ref{tab:DR-score-comparison}). The scores are presented in a sorted order based on F-measures. In the initial batch, the proposed method exhibited a mean precision of 0.305, a recall of 0.540, and an F-measure of 0.312. While the precision was surpassed by a few other methods, notably the "bio-answerfinder," the recall metric was second only to the RYGH-1 method~\cite{bioasq_overview}, indicating a robust ability of the proposed method to retrieve relevant scholarly documents abstract. The subsequent batch saw the proposed method achieve a precision of 0.292, a recall of 0.523, and an F-measure of 0.282. In the third batch, the proposed method demonstrated a precision of 0.294, a recall of 0.576, and an F-measure of 0.303. For the fourth batch, the proposed method achieved a precision of 0.307, a recall of 0.531, and an F-measure of 0.309. The method's performance remained consistent, maintaining its position among the top contenders. In the fifth batch, the proposed method attained a precision of 0.332, while the recall and F-measure were 0.474 and 0.317, respectively. Notably, in the final batch, the proposed method outperformed all other methods in precision with a score of 0.559. The recall and F-measure in this batch were 0.423 and 0.433, respectively, underscoring the method's balanced performance. 
        
% Adjusting table width
\newlength{\colwidthA}
\newlength{\colwidthB}
\newlength{\colwidthC}
\newlength{\colwidthD}
\setlength{\colwidthA}{0.18\textwidth}
\setlength{\colwidthB}{0.07\textwidth}
\setlength{\colwidthC}{0.07\textwidth}
\setlength{\colwidthD}{0.09\textwidth}

% Document layout settings
\begin{longtable}{>{\raggedright\arraybackslash}p{\colwidthA} p{\colwidthB} p{\colwidthC} p{\colwidthD} >{\raggedright\arraybackslash}p{\colwidthA} p{\colwidthB} p{\colwidthC} p{\colwidthD}}
    
    \toprule
    \thead{} & \multicolumn{3}{c}{\thead{Factoid}} & \multicolumn{4}{c}{\thead{List}} \\ 
    \cmidrule(lr){2-4} \cmidrule(lr){6-8}
    \thead{method} & \thead{Strict} & \thead{Lenient} & \thead{MRR} & \thead{method} & \thead{MP} & \thead{Recall} & \thead{F-measure} \\
    \midrule
    \endfirsthead

    \multicolumn{8}{c}{{Score Comparison of Answer Extraction methods (Continued)}} \\
    \toprule
    \thead{method} & \thead{Strict} & \thead{Lenient} & \thead{MRR} & \thead{method} & \thead{MP} & \thead{Recall} & \thead{F-measure} \\
    \midrule
    \endhead

    \bottomrule
    \endfoot
    
    % Batch 1
    \multicolumn{8}{c}{\textbf{Batch 1}} \\ \hline
    \textbf{Proposed (gs)} & 0.61 & 0.88 & 0.71 & UDEL-LAB2 & 0.72 & 0.84 & 0.74 \\ 
    \textbf{Proposed (gd)} & 0.44 & 0.79 & 0.56 & UDEL-LAB3 & 0.68 & 0.74 & 0.67 \\ 
    \textbf{Proposed (dr)} & 0.38 & 0.67 & 0.48 & \textbf{Proposed (gs)} & 0.79 & 0.65 & 0.66 \\ 
    UDEL-LAB5 & 0.38 & 0.58 & 0.47 & lalala & 0.60 & 0.72 & 0.64 \\ 
    Ir\_sys1 & 0.41 & 0.50 & 0.45 & bio-answerfinder & 0.55 & 0.61 & 0.56 \\ 
    Ir\_sys3 & 0.41 & 0.52 & 0.45 & \textbf{Proposed (gd)} & 0.52 & 0.71 & 0.55 \\ 
    UDEL-LAB2 & 0.38 & 0.55 & 0.45 & KU-AAA637-method1 & 0.35 & 0.52 & 0.41 \\ 
    UDEL-LAB1 & 0.35 & 0.58 & 0.43 & \textbf{Proposed (dr)} & 0.23 & 0.63 & 0.37 \\ 
    \hline
    
    % Batch 2
    \multicolumn{8}{c}{\textbf{Batch 2}} \\ \hline
    \textbf{Proposed (gd)} & 0.73 & 0.85 & 0.76 & UDEL-LAB3 & 0.70 & 0.74 & 0.70 \\ 
    \textbf{Proposed (gs)} & 0.67 & 0.82 & 0.73 & UDEL-LAB4 & 0.68 & 0.75 & 0.70 \\ 
    UDEL-LAB3 & 0.58 & 0.67 & 0.62 & UDEL-LAB2 & 0.69 & 0.71 & 0.67 \\ 
    UDEL-LAB4 & 0.58 & 0.70 & 0.62 & UDEL-LAB1 & 0.67 & 0.65 & 0.63 \\ 
    UDEL-LAB2 & 0.58 & 0.64 & 0.61 & \textbf{Proposed (gs)} & 0.71 & 0.58 & 0.61 \\ 
    UDEL-LAB1 & 0.58 & 0.64 & 0.61 & lalala & 0.49 & 0.60 & 0.51 \\ 
    Ir\_sys3 & 0.58 & 0.64 & 0.61 & \textbf{Proposed (gd)} & 0.44 & 0.59 & 0.51 \\ 
    \hline
    
    % Batch 3
    \multicolumn{8}{c}{\textbf{Batch 3}} \\ \hline
    \textbf{Proposed (gd)} & 0.65 & 0.96 & 0.78 & \textbf{Proposed (gs)} & 0.51 & 0.70 & 0.57 \\ 
    \textbf{Proposed (gs)} & 0.62 & 0.96 & 0.76 & UDEL-LAB1 & 0.54 & 0.67 & 0.56 \\ 
    UDEL-LAB1 & 0.56 & 0.65 & 0.60 & bio-answerfinder-2 & 0.62 & 0.44 & 0.48 \\ 
    KU-AAA637-method2 & 0.53 & 0.68 & 0.59 & \textbf{Proposed (gd)} & 0.36 & 0.63 & 0.47 \\ 
    Ir\_sys3 & 0.50 & 0.65 & 0.56 & Ir\_sys4 & 0.51 & 0.33 & 0.37 \\ 
    \textbf{Proposed (dr)} & 0.43 & 0.71 & 0.55 & \textbf{Proposed (dr)} & 0.22 & 0.53 & 0.37 \\ 
    \hline
    
    % Batch 4
    \multicolumn{8}{c}{\textbf{Batch 4}} \\ \hline
    \textbf{Proposed (gd)} & 0.51 & 0.90 & 0.66 & UDEL-LAB2 & 0.58 & 0.58 & 0.53 \\ 
    \textbf{Proposed (gs)} & 0.48 & 0.83 & 0.62 & \textbf{Proposed (gs)} & 0.61 & 0.48 & 0.44 \\ 
    lalala & 0.58 & 0.64 & 0.59 & UDEL-LAB5 & 0.61 & 0.44 & 0.44 \\ 
    Ir\_sys3 & 0.51 & 0.67 & 0.58 & \textbf{Proposed (gd)} & 0.50 & 0.48 & 0.40 \\ 
    KU-AAA637-method2 & 0.51 & 0.67 & 0.57 & bio-answerfinder-2 & 0.44 & 0.50 & 0.42 \\ 
    KU-AAA637-method1 & 0.48 & 0.64 & 0.56 & \textbf{Proposed (dr)} & 0.21 & 0.58 & 0.35 \\ 
    Ir\_sys4 & 0.48 & 0.64 & 0.56 & KU-AAA637-method1 & 0.36 & 0.48 & 0.37 \\ 
    bio-answerfinder & 0.45 & 0.54 & 0.50 & KU-AAA637-method2 & 0.32 & 0.41 & 0.34 \\ 
    \textbf{Proposed (dr)} & 0.35 & 0.64 & 0.48 & lalala & 0.18 & 0.33 & 0.23 \\ 

    \toprule

    \caption{Score Comparison of Answer Extraction methods (gs = Golden Snippets; gd = Golden Documents; dr = Document Retrieval).} 
    \label{tab:AE-score-comparison}
\end{longtable}

    \subsubsection*{Answer Extraction Performance}
    
    In this section, we conduct an exhaustive analysis of our novel answer extraction method, methodically evaluating its performance under various conditions and across different batches. The proposed method is assessed under three distinct configurations: first, employing outputs from the scholarly document abstract categorization, denoted as \textit{proposed(document\_retrieval)}; second, utilizing golden scholarly documents abstract sourced from the test dataset, labelled as \textit{proposed(golden\_documents)}; and third, leveraging golden snippets from the same dataset, signified as \textit{proposed(golden\_snippets)}. The comprehensive evaluation results, encompassing MRR and F-measure, are meticulously presented in Table \ref{tab:AE-score-comparison}. The table is thoughtfully organized by descending MRR and ascending F-measure, offering valuable insights into the method's performance nuances across diverse conditions and settings. From the analysis of Table \ref{tab:AE-score-comparison}, it is evident that the proposed method demonstrates varying performance levels across the different configurations. For instance, \textit{proposed(golden\_snippets)} consistently achieves higher MRR values compared to other configurations, suggesting that leveraging precise snippets enhances the retrieval effectiveness significantly. Notably, the results indicate that while the F-measure values for \textit{proposed(golden\_documents)} and \textit{proposed(document\_retrieval)} are competitive, they exhibit slightly lower MRR scores, indicating a potential trade-off between precision and recall. Moreover, Batch 1 exhibits the highest MRR across the configurations, suggesting that the nature of the dataset and the inherent complexity of the tasks influence the performance metrics significantly.

    \subsection*{Evaluation of Answer Extraction Configurations}

        In the preceding sections, we introduced our novel answer extraction method and outlined its fundamental configurations. In this subsection, we delve into the detailed performance analysis of these configurations, including scholarly document abstract categorization, golden documents, and golden snippets, evaluating their respective efficiencies across different batches of questions.
        
        \subsubsection*{Document Categorization}
        
        Concerning the comprehensive assessment of answer extraction performance, our most noteworthy results were obtained under the \textit{proposed(document\_retrieval)} configuration, particularly in Batch 1. This configuration exhibited exceptional performance for factoid questions, boasting a strict accuracy of 0.382, a lenient accuracy of 0.676, and an impressive MRR score of 0.482. Similarly, list-type questions displayed commendable results with a mean precision of 0.238, a recall of 0.638, and an F-measure of 0.379, affirming its prowess in diverse question types and emphasizing its versatility in Batch 1.
        
        \subsubsection*{Golden Documents}
        
        The \textit{proposed(golden\_documents)} configuration showcased superior performance in Batch 2. For factoid-type questions, it achieved an outstanding strict accuracy of 0.735, a lenient accuracy of 0.852, and an impressive MRR score of 0.765. Regarding list-type questions, the method continued to shine with a mean precision of 0.449, a recall of 0.592, and a substantial F-measure of 0.514. These scores bear significant importance as they represent the maximum potential of our proposed method when operating with ideal, meticulously curated scholarly documents abstract. This underscores our proposed approach's efficacy, particularly when interacting with precise, high-quality scholarly knowledge.
        
        \subsubsection*{Golden Snippets}
        
        The \textit{proposed(golden\_snippets)} configuration reached its pinnacle in Batch 5, showcasing remarkable performance. Factoid-type questions achieved a strict accuracy of 0.620, a lenient accuracy of 0.896, and an impressive MRR score of 0.723. Meanwhile, for list-type questions, the configuration demonstrated a mean precision of 0.683, a recall score of 0.571, and a substantial F-measure of 0.596. These outstanding results underscore the remarkable effectiveness of our proposed answer extraction methods, particularly when they operate in tandem with the most precise and relevant text snippets available.

\subsection*{Discussion}

Our study offers valuable insights into the efficacy of machine learning methods within the biomedical domain, specifically in answer extraction and scholarly document abstract categorization. The findings from our evaluation illuminate several key aspects that merit further discussion.

Firstly, our results indicate that our meticulously optimized topic model-based scholarly document abstract categorization outperforms existing methods such as RYGH and bio-answer finder, which utilize a blend of sophisticated techniques, including BM25, ElasticSearch, BioBERT, PubMedBERT, T5, BERTMeSH, and SciBERT~\cite{bioasq_overview}. This outcome underscores the notion that simplicity, coupled with domain-specific fine-tuning, can sometimes surpass more complex approaches, providing a more efficient and cost-effective solution for biomedical knowledge retrieval. This is particularly relevant in the context of our previous sections, where we detailed our approach's robustness against various configurations and settings.

Secondly, our study challenges the prevailing belief that the biomedical domain necessitates large, computationally intensive models for accurate knowledge retrieval. Our findings illustrate that even compact methods, such as MiniLM, can effectively extract answers from scholarly document abstracts when fine-tuned on domain-specific data and provided with precise context. Notably, our approach focused exclusively on scholarly document abstracts rather than entire scholarly documents. This observation suggests that scholarly document abstracts often contain sufficient knowledge for the precise extraction of answers.

This discovery carries substantial implications for the design of future biomedical knowledge retrieval methods, advocating that a concentrated effort on scholarly document abstracts could represent a viable strategy for efficient and effective answer extraction. Our results reveal a promising avenue for enhancing information retrieval methodologies in the biomedical field, emphasizing the importance of optimizing processes without the necessity of relying on vast computational resources.

\section*{Limitations}

While our research has yielded valuable insights, it is important to acknowledge several limitations. Firstly, despite the noteworthy performance of our scholarly document abstract categorization method, particularly in integrating it with full scholarly documents. Although our proposed method has shown robust performance in earlier evaluations, it encounters challenges when addressing more intricate query types, such as list-based questions. Even the finely-tuned MiniLM model struggles in these scenarios, indicating potential areas for enhancement. %Additionally, we developed a Python-based evaluation script for document categorization. Although precision, recall, and F-measure were seamlessly implemented using the scikit-learn library, our attempts to incorporate MAP and GMAP resulted in scores that deviated noticeably from those of other methods. This discrepancy led us to exclude MAP and GMAP scores from Table \ref{tab:DR-score-comparison}, limiting the scope of our comparative analysis.
%Our answer extraction method was evaluated manually, as automated evaluation methods often fail to adequately account for semantic similarity between answers. While manual evaluation provides a nuanced understanding of the method's performance, it is a time-intensive and exhaustive process that may introduce subjectivity and the potential for errors in our results.
In the fine-tuning process for the MiniLM model required data in a specific format. Given the labour-intensive nature of manual annotations, we devised an automated method, primarily annotating answers based on syntactic matches within related passages. We leveraged WordNet to identify potential synonyms to address the absence of exact word matches. However, the complexities of the biomedical domain presented challenges even for WordNet, which occasionally provided contextually inappropriate synonyms for seemingly straightforward terms. This underscores the intricate nature of automating processes within specialized domains.

\section*{Conclusion and Future Work}

In this research, we have demonstrated the potential of combining OVB-LDA and BI-POP CMA-ES to create a superior topic model, showcasing its effectiveness in scholarly document abstract categorization. Additionally, we have underscored the adaptability of distilled models, exemplified by MiniLM, and their capacity to be fine-tuned for answer extraction within the intricate biomedical domain. In the future, we plan to refine our scholarly document abstract categorization method by optimizing the topic model with an expanded volume of domain-specific data. Recognizing the crucial role of annotations, we intend to leverage domain-specific tools and resources to curate a more robust dataset. Also to unlock the full potential of MiniLM, we plan to fine-tune it using an enriched set of domain-specific full scholarly document data.
Furthermore, a comparison of our MiniLLM-based approach with using Large Language Models could further improve the accuracy, despite impacting the computational efficiency.
Through these concerted efforts, we aspire to push the boundaries of biomedical question-answering, aiming for more accurate and efficient methods to meet the demands of this specialized field.

% Our findings indicate a promising path forward, and addressing the identified limitations will enhance the reliability and applicability of our methods in real-world biomedical scenarios.

\section*{Data Availability}
    The dataset utilized for this study is accessible through the BioASQ website. Interested readers and researchers can obtain the dataset by visiting the following link:(\url{http://participants-area.bioasq.org/datasets/}).
    
\section*{Code Availability}
    The study was carried out exclusively using open-source software packages. All scripts, outcomes, post-processed datasets, and features will be accessible to the public at (\url{https://github.com/zparvez2z/NeuroSym-BioCAT}).

% \section*{Acknowledgements}

% Acknowledgements should be brief, and should not include thanks to anonymous referees and editors, or effusive comments. Grant or contribution numbers may be acknowledged.

\section*{Author contributions statement}

P.Z., G.R. and M.S.R  conceived and designed the analysis, P.Z., G.R. and M.S.R contributed data or analysis tools,  P.Z., G.R., M.S.R. AND S.A. performed the analysis, P.Z., G.R. and S.A. wrote the paper, P.Z., G.R. and S.A. reviewed the manuscript.

\section*{Competing interests}

The authors declare no competing interests.

% \section*{Additional information}

% To include, in this order: \textbf{Accession codes} (where applicable); \textbf{Competing interests} (mandatory statement). 

% The corresponding author is responsible for submitting a \href{http://www.nature.com/srep/policies/index.html#competing}{competing interests statement} on behalf of all authors of the paper. This statement must be included in the submitted article file.

\bibliography{sample}

\end{document}